\documentclass{article}
\usepackage{amsmath,epsfig}
\usepackage[preprint]{spconfa4}
\usepackage{booktabs}
\usepackage{url}

\copyrightnotice{978-1-6654-3864-3/21/\$31.00~\copyright 2021 IEEE}

\let\OLDthebibliography\thebibliography
\renewcommand\thebibliography[1]{
  \OLDthebibliography{#1}
  \setlength{\parskip}{0pt}
  \setlength{\itemsep}{0pt plus 0.3ex}
}

\begin{document}\sloppy

\def\x{{\mathbf x}}
\def\L{{\cal L}}
\def\lyu{\textcolor{black}}

\title{CORE-Text: Improving Scene Text Detection with\\ Contrastive Relational Reasoning }
%
\name{Jingyang Lin$^{\ast}$$^{\dagger}$, Yingwei Pan$^{\ddagger}$, Rongfeng Lai$^{\ddagger}$, Xuehang Yang$^{\ddagger}$, Hongyang Chao$^{\ast}$$^{\dagger}$, and Ting Yao$^{\ddagger}$}
\address{$^{\ast}$Sun Yat-sen University, Guangzhou, China\\
         $^{\dagger}$The Key Laboratory of Machine Intelligence and Advanced Computing (Sun Yat-sen University),\\ Ministry of Education, Guangzhou, P. R. China\\
         $^{\ddagger}$JD AI Research, Beijing, China\\
		 {\tt\small \{yung.linjy, panyw.ustc\}@gmail.com, lairf@foxmail.com, yangxuehang@jd.com,}\\
         {\tt\small isschhy@mail.sysu.edu.cn, tingyao.ustc@gmail.com}
}

\maketitle

\begin{abstract}
Localizing text instances in natural scenes is regarded as a fundamental challenge in computer vision.
Nevertheless, owing to the extremely varied aspect ratios and scales of text instances in real scenes, most conventional text detectors suffer from the {\em sub-text} problem that only localizes the fragments of text instance (i.e., {\em sub-texts}).
In this work, we quantitatively analyze the {\em sub-text} problem and present a simple yet effective design, \textbf{CO}ntrastive \textbf{RE}lation (\textbf{CORE}) module, to mitigate that issue.
CORE first leverages a vanilla relation block to model the relations among all text proposals (\emph{sub-texts} of multiple text instances) and further enhances relational reasoning via instance-level sub-text discrimination in a contrastive manner.
Such way naturally learns instance-aware representations of text proposals and thus facilitates scene text detection.
We integrate the CORE module into a two-stage text detector of Mask R-CNN and devise our text detector CORE-Text. Extensive experiments on four benchmarks demonstrate the superiority of CORE-Text. Code is available: \url{https://github.com/jylins/CORE-Text}.

\end{abstract}

\begin{keywords}
Scene text detection, Relational reasoning, Contrastive learning
\end{keywords}
\vspace{-1em}

\renewcommand{\thefootnote}{}
\footnote{\noindent This work was performed at JD AI Research, and was partially supported by NSF of China under Grant 61672548, U1611461.}

\section{Introduction}\label{sec:intro}

\begin{figure}[t]
\vspace{-0.28in}
\begin{minipage}[b]{0.36\linewidth}
  \centerline{\epsfig{figure=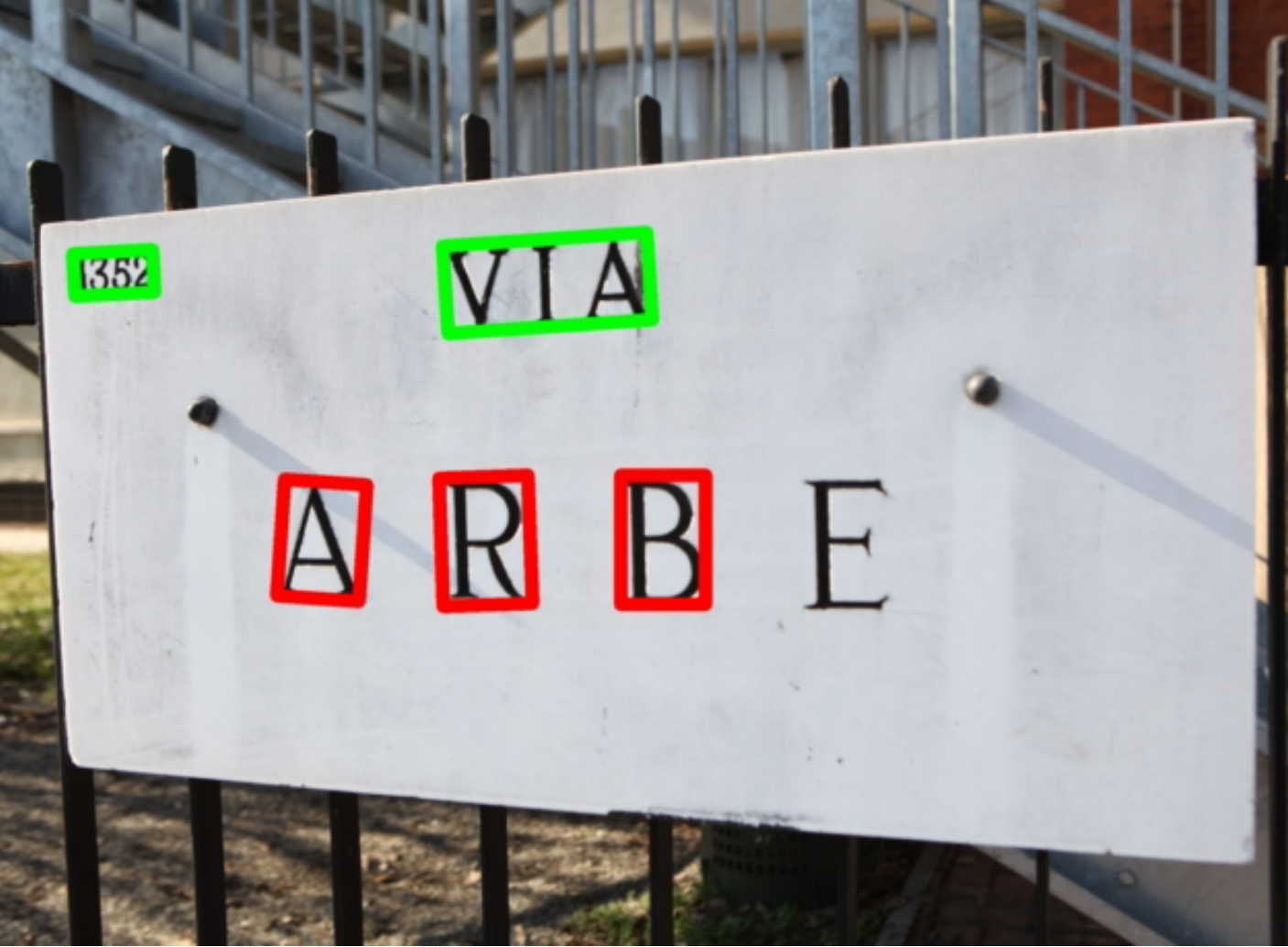,height=2.4cm, width=3cm}}
\end{minipage}
\vspace{-0.05in}
\begin{minipage}[b]{0.234\linewidth}
  \centerline{\epsfig{figure=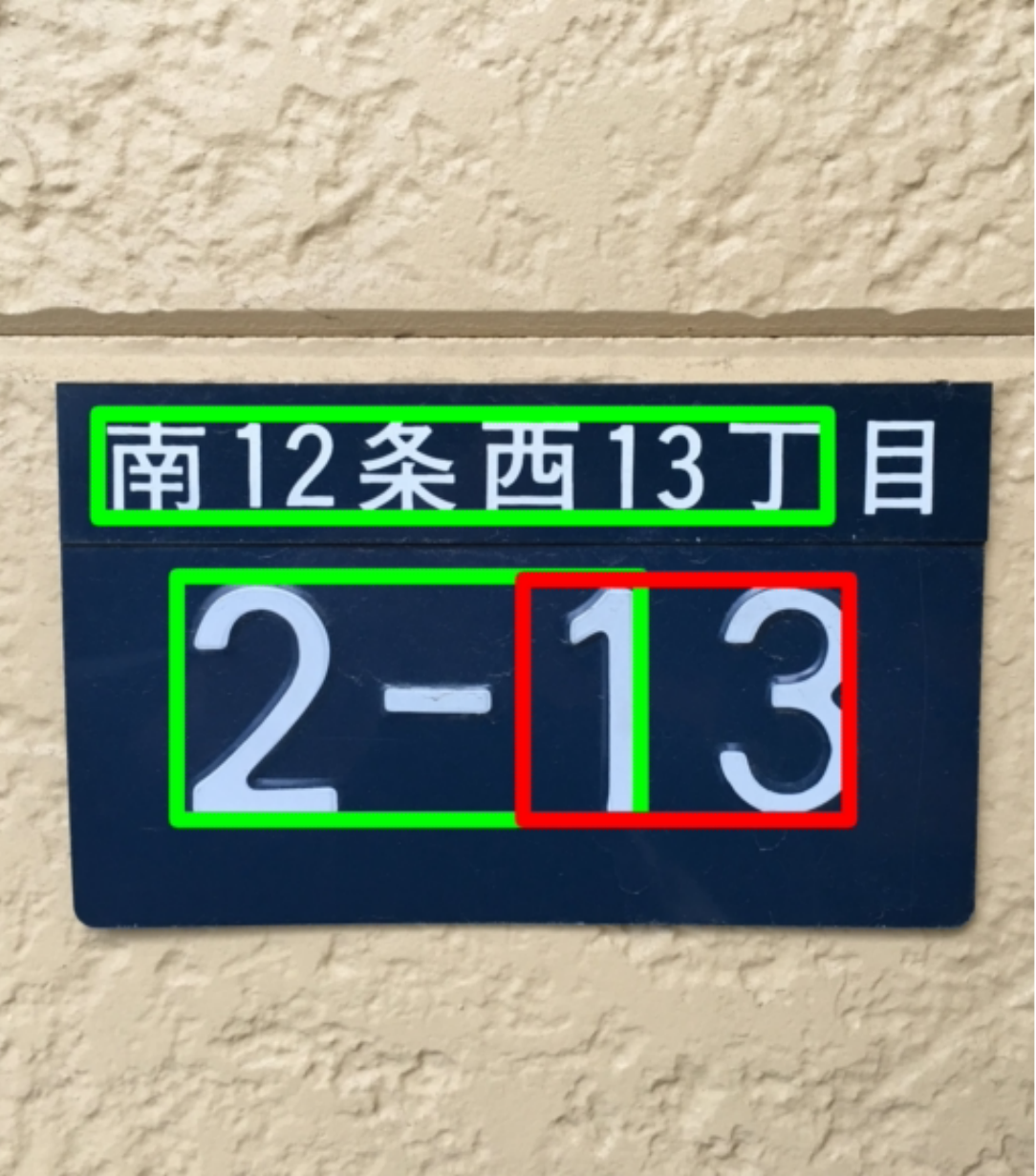,height=2.4cm, width=2.2cm}}
\end{minipage}
\begin{minipage}[b]{0.387\linewidth}
  \centerline{\epsfig{figure=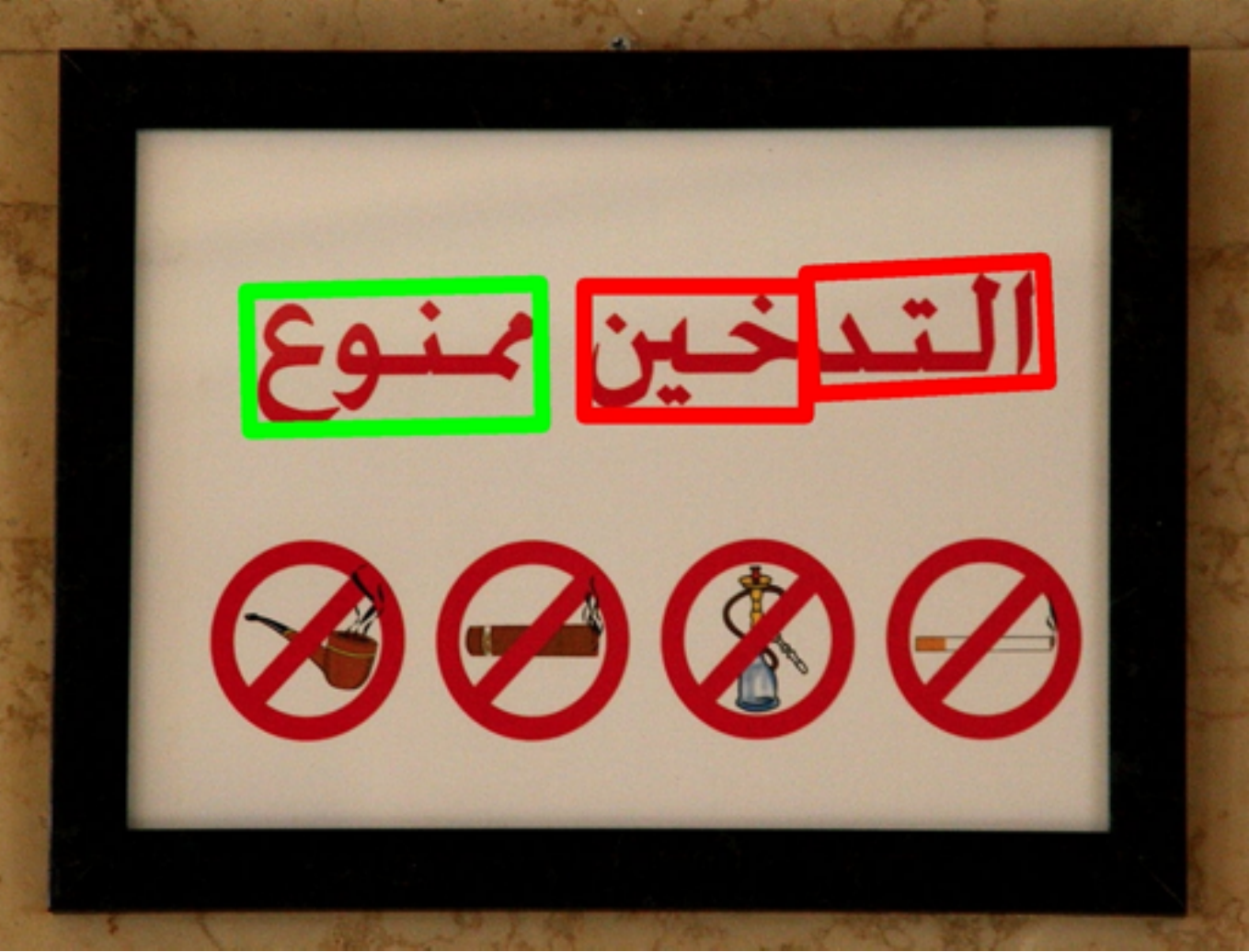,height=2.4cm,width=3.24cm}}
\end{minipage}
\centerline{\small (a) Results of generic object detector (Mask R-CNN)}
\begin{minipage}[b]{0.36\linewidth}
  \centerline{\epsfig{figure=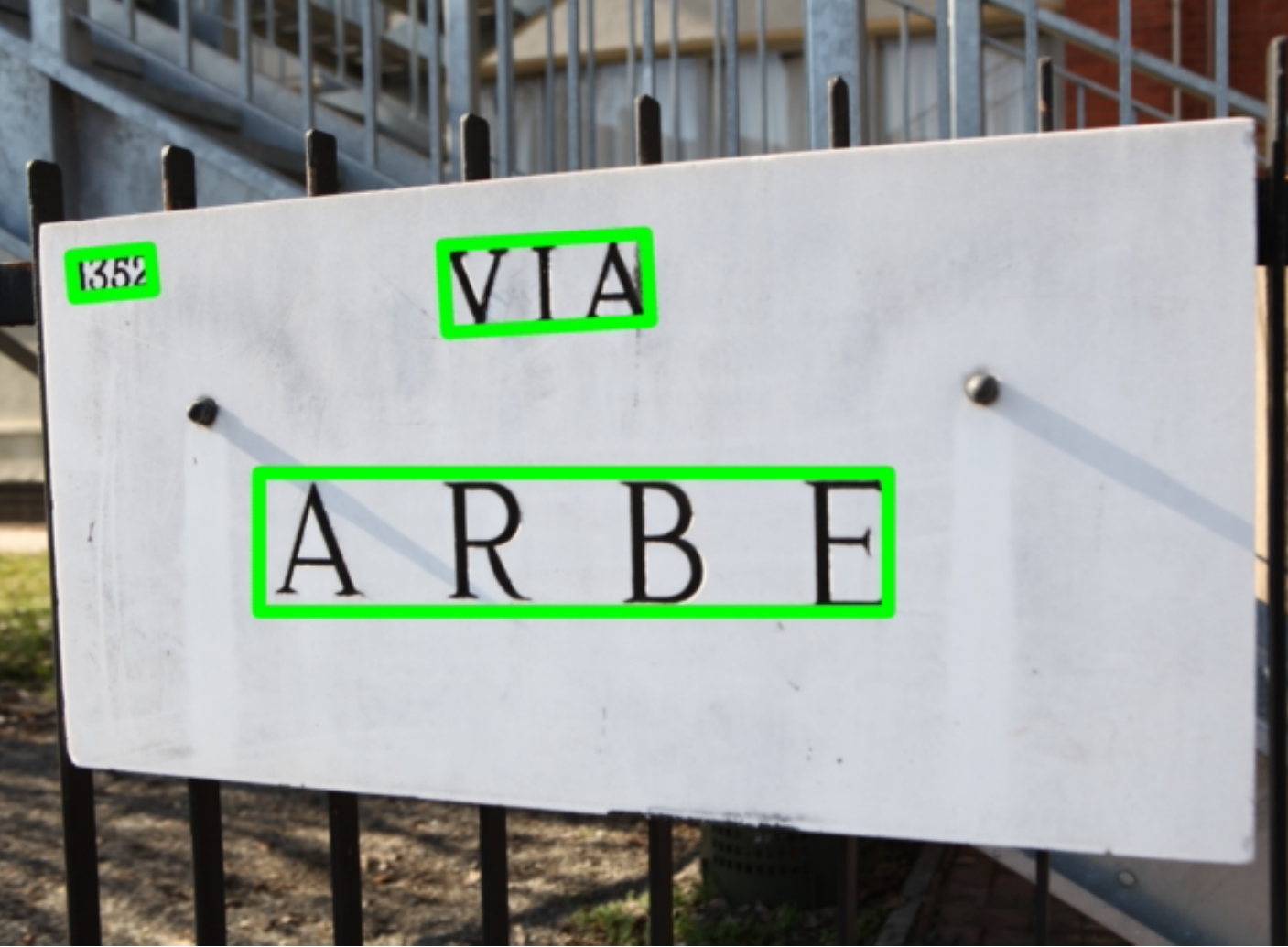,height=2.4cm, width=3cm}}
\end{minipage}
\vspace{-0.05in}
\begin{minipage}[b]{0.234\linewidth}
  \centerline{\epsfig{figure=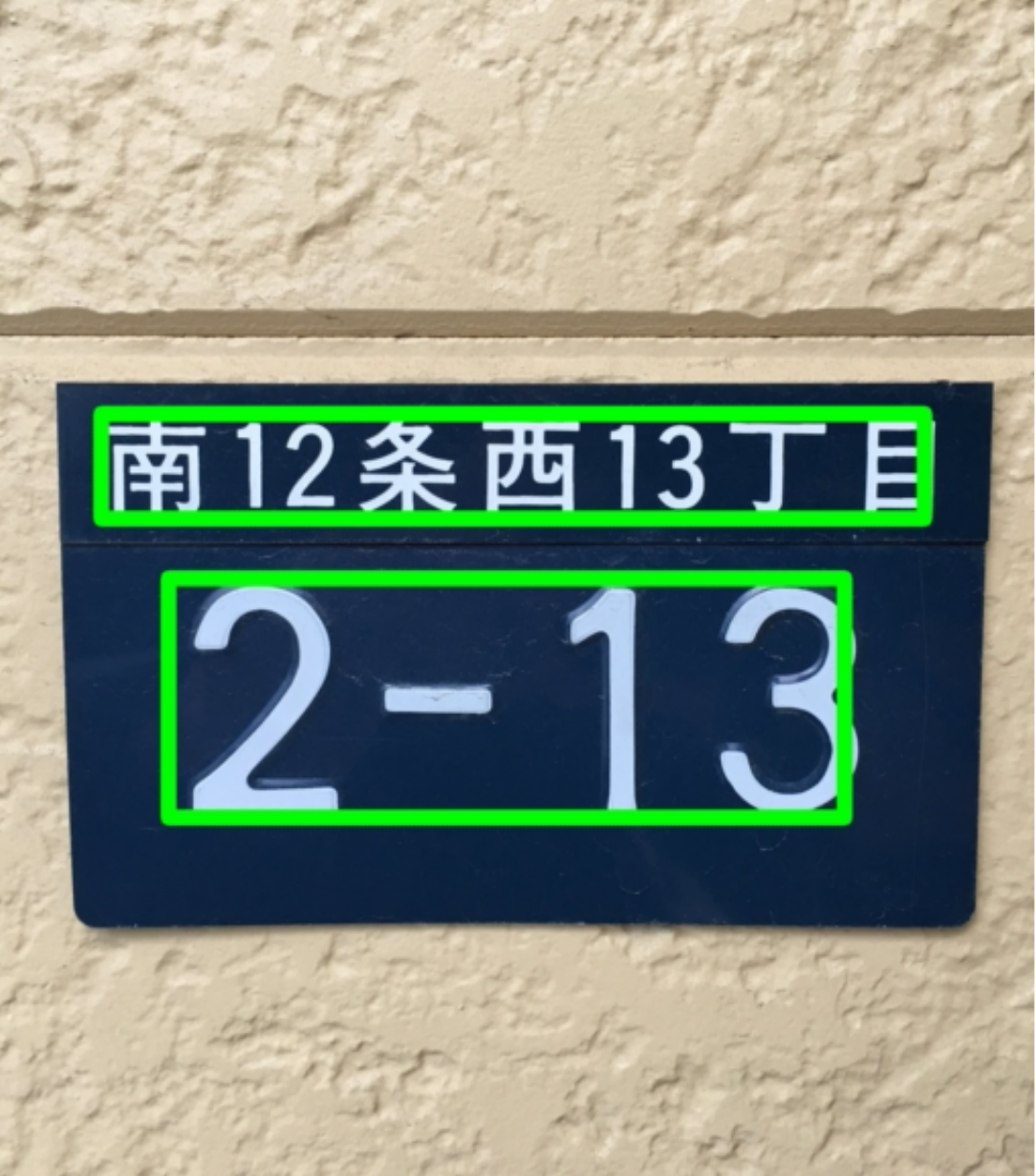,height=2.4cm, width=2.2cm}}
\end{minipage}
\begin{minipage}[b]{0.387 \linewidth}
  \centerline{\epsfig{figure=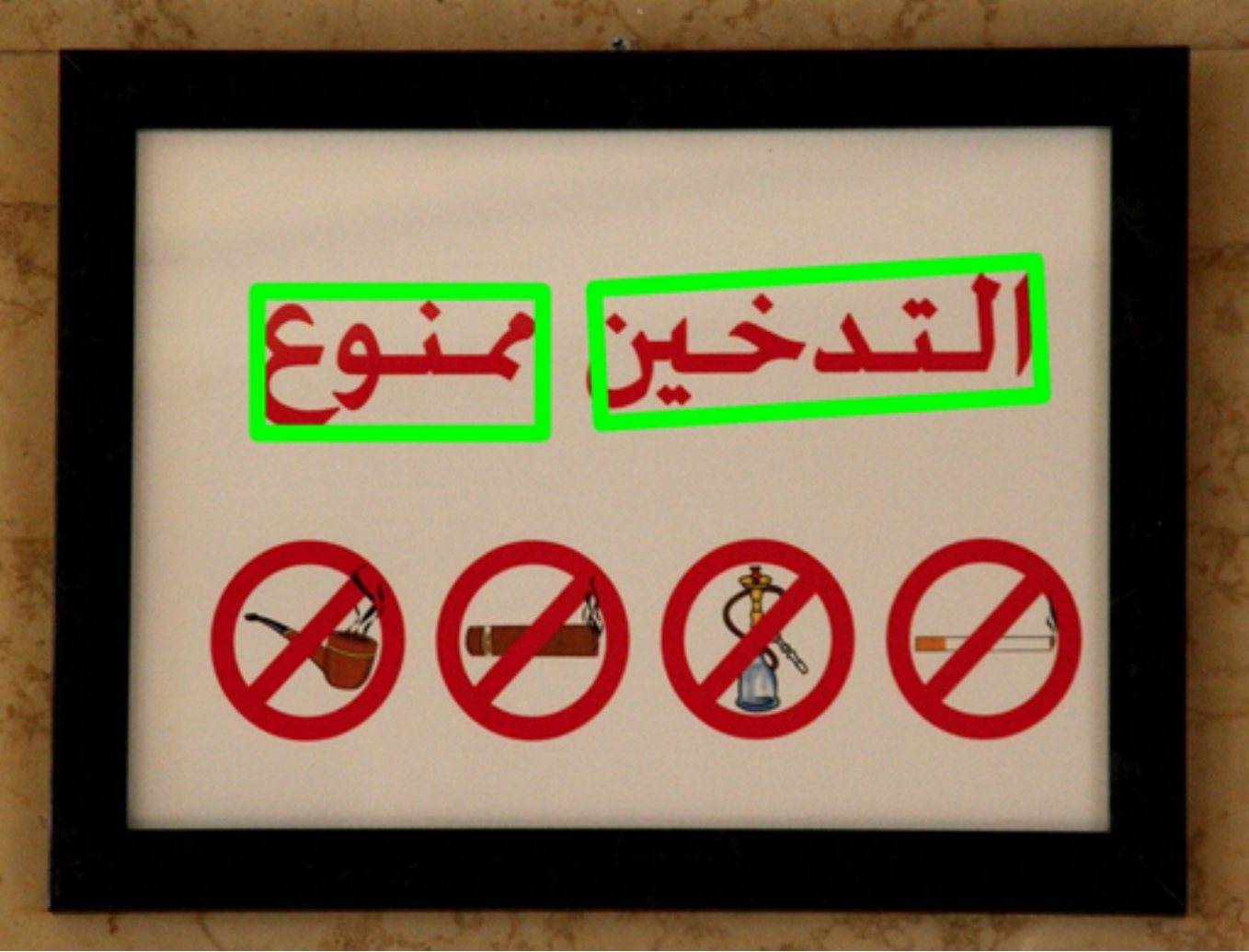,height=2.4cm,width=3.24cm}}
\end{minipage}
\centerline{\small (b) Results of our CORE-Text}
\vspace{-0.3in}
\caption{\small Scene text detection on three images by (a) directly applying generic object detector (Mask R-CNN) and (b) utilizing CORE-Text in this work. Red box: {\em sub-text} detection that merely detects sub-regions of \emph{full-text} instance; Green box: {\em full-text} detection.}
\vspace{-0.25in}
\label{fig.intro}
\end{figure}

Scene text detection, which is known as localizing text instances in natural scene images, is a profound challenge in both computer vision and deep learning communities. Practical automatic scene text detection systems have a great potential impact for numerous applications, e.g., document analysis, industrial automatic, and autonomous driving.
The recent development of deep learning techniques for generic object detection \cite{det:fpn,det:faster-rcnn} and segmentation~\cite{ins:mask-rcnn,segm:fcn} has successfully pushed the limits of scene text detection, leading to a surge of deep text detector \cite{text:ctpn, text:seglink,  text:bootstrap, text:drrg,text:pmtd} that follow the typical region proposal-based detection paradigm.
Nevertheless, considering that the aspect ratios and scales of text instances often suffer from more variation than those of generic objects, directly applying generic object detectors will inevitably result in broken detections of text instances \cite{text:seglink,text:bootstrap}. Taking the text detection results in Figure \ref{fig.intro} (a) as an example, the generic object detector (Mask R-CNN \cite{ins:mask-rcnn}) fails to accurately localize the whole text instances (i.e., {\em full-text}) and only detects them as multiple text fragments (i.e., {\em sub-text}), especially when the aspect ratios of text instances are large. These facts motivate the exploration of contextual information among {\em sub-texts} to alleviate this {\em sub-text} problem.

In the literature, there have been a series of innovations being proposed to improve scene text detection through exploiting contextual information among {\em sub-texts} to associate the {\em sub-texts} belonging to the same text instance, e.g., segment linking \cite{text:seglink} or link merging over local graphs \cite{text:drrg}.
Nevertheless, most of them solve the {\em sub-text} problem in a two-phase manner, i.e., first localizing {\em sub-texts} of multiple text instances in an image and then grouping the {\em sub-texts} of the same instance.
Such way may break the integration between localizing and associating {\em sub-texts} of the same text instance, resulting in a sub-optimal solution.
Moreover, though these methods have demonstrated performance gains in detection accuracy by addressing the {\em sub-text} problem, it is still unclear to what extent the {\em sub-text} problem affects the overall performance of text detector for scene text detection task.

In this work, we engage in solving the {\em sub-text} problem in scene text detection. First, we quantitatively analyze the frequency of the {\em sub-text} problem for a generic object detector (Mask R-CNN) on the benchmark (e.g., ICDAR 2017 MLT) and provide the performance upper-bound by fully eliminating the negative effect of {\em sub-texts}. Surprisingly, we find that the {\em sub-text} problem accounts for a large proportion of bad cases in existing benchmark, and a significant performance boost ($\geq 6$\% in Hmean metric) is attained when the {\em sub-text} problem is fully addressed.

Moreover, by consolidating the idea of unifying both localization and association of text proposals (containing {\em sub-texts}), we present a novel COntrastive RElation (CORE) module to mitigate the {\em sub-text} problem in scene text detection task. Technically, the multi-scale text proposals, i.e., a group of {\em sub-texts} and {\em full-texts} derived from multiple text instances in an input image, are first produced via Region Proposal Networks (RPN). Next, we leverage a vanilla relation block~\cite{det:rel-net} to perform relational reasoning among all text proposals. The relational reasoning is further guided with instance-wise contrastive objective, that pursues instance-level {\em sub-text} discrimination in a contrastive manner. This design pursues the learning of instance-aware representations of text proposals through jointly relational reasoning and text instance identification, and thus facilitates the localization and classification of text proposals. Our CORE module could be regarded as a general text proposal refiner and is readily pluggable to any two-stage text detector. We name the whole architecture of text detector (Mask R-CNN here) with CORE module as CORE-Text, and empirically demonstrate that CORE-Text could better mitigate the {\em sub-text} problem and obtain encouraging detection results as illustrated in Figure \ref{fig.intro} (b).

\section{Related Work}

{\bf Scene Text Detection.} Recent progress on this task has evolved through two paradigms: segmentation-based~\cite{text:psenet, text:db} and proposal-based methods~\cite{text:ctpn, text:seglink, text:bootstrap, text:drrg,text:pmtd}. The primary challenge of the former is to distinguish text instances from pixel perspective. The latter may suffer from various aspect ratios and scales of scene texts, and result in the {\em sub-text} problem. Several existing works~\cite{text:ctpn, text:seglink, text:bootstrap, text:drrg} mitigate the {\em sub-text} problem in a two-phase way (i.e. localizing and grouping {\em sub-texts}), which may break the interaction between localizing and associating {\em sub-text}, and lead to a sub-optimal solution. In contrast, we unify both localization and association of text proposals, without any additional grouping post-process.

{\bf Relational Reasoning.} There has been strong evidences on the use of relational reasoning to support various tasks, e.g., object detection~\cite{det:rel-net,cai2019exploring,deng2019relation,det:nonlocal}, feature learning \cite{pan2016learning}, vision-language \cite{pan2020x,yao2018exploring}. For example, \cite{det:nonlocal} plugs non-local operation into the conventional CNN to enable the pixel-level relational interaction within feature maps, and \cite{det:rel-net} presents an object relation module to model the relations of regions via the interaction among appearance features and geometry.

The CORE module in our work is also a type of object-to-object relational reasoning. Unlike~\cite{det:rel-net} that is developed for generic object detection, ours goes beyond the self-supervised exploration of contextual information among proposals and aims to additionally guide relational reasoning with instance-wise contrastive objective to mitigate \emph{sub-text} problem. Such way naturally unifies both localization and association of text proposals (consisting of \emph{sub-texts} and \emph{full-texts}), and thus facilitates scene text detection.

\section{Sub-text Problem}
\label{sec:prob}

The {\em sub-text} problem refers to the broken detection results in scene text detection task, where a text instance is detected as multiple text fragments ({\em sub-texts}). Though the unsatisfactory results caused by {\em sub-texts} have been mentioned in several existing works~\cite{text:seglink,text:bootstrap}, the problem of how the {\em sub-texts} affect the overall performance of text detector is not yet fully understood in the literature. In this section, we look into this problem, and provide a detailed quantitative analysis of {\em sub-text} problem, including the concrete definition of {\em sub-text}, the frequency of {\em sub-text} problem in bad cases, and the performance upper-bound \footnote{The performances are reported on ICDAR 2017 MLT {\em val} set, and the detail experiment setting can be referred in Section~\ref{sec.analysis}.} if the {\em sub-text} problem is solved.

\begin{figure}[t]
\vspace{-0.12in}
\begin{minipage}[b]{0.48\linewidth}
  \centerline{\epsfig{figure=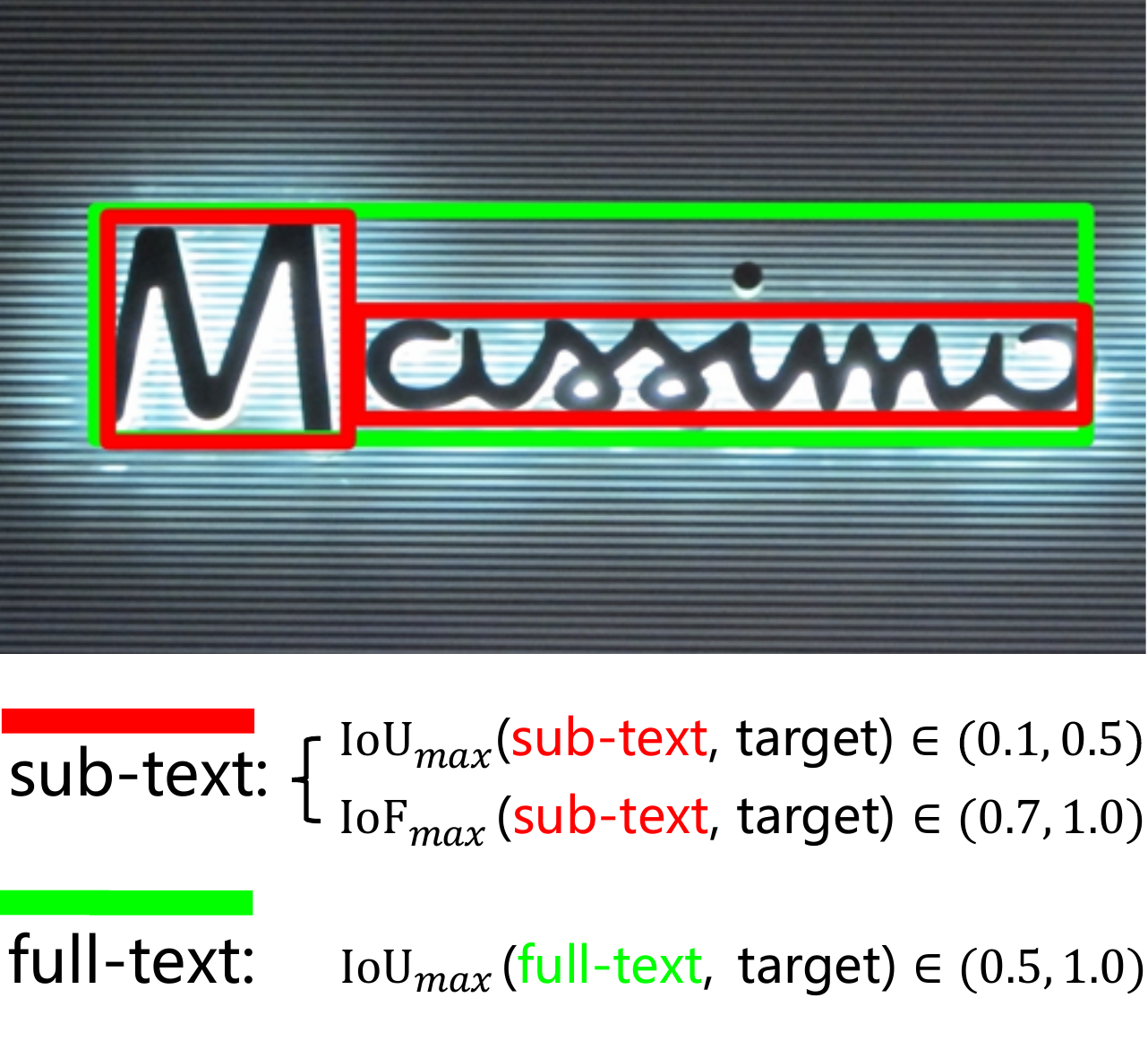,width=4cm}}
  \vspace{-0.1in}
  \centerline{\small (a) {\em Sub-text} \& {\em full-text} definition}
\end{minipage}
\hfill
\begin{minipage}[b]{.512\linewidth}
  \leftline{\epsfig{figure=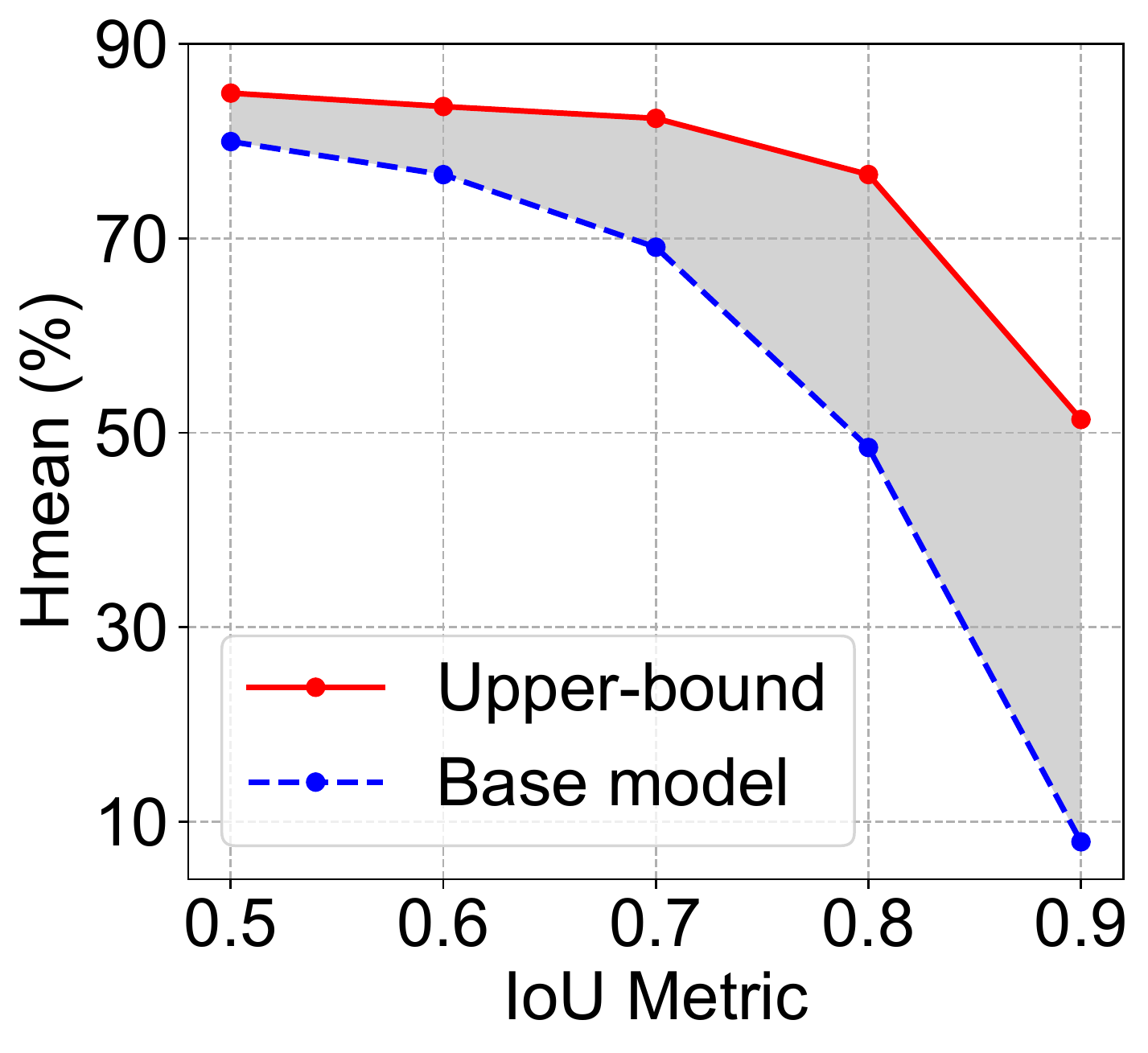,height=3.8cm}}
  \vspace{-0.1in}
  \centerline{\small (b) Performance upper-bound}
\end{minipage}
\vspace{-0.3in}
\caption{\small Quantitative analysis of {\em sub-text} problem: (a) the definition of {\em sub-text} and {\em full-text}; (b) the performance upper-bound when {\em sub-text} problem is fully addressed.}
\vspace{-0.22in}
\label{fig:sub-text}
\end{figure}

{\bf Sub-text Definition.} To formalize this problem, we first present the concrete definition of {\em sub-text} and {\em full-text} conditioned on the relative position between the detection proposal and ground truth (Figure~\ref{fig:sub-text}(a)). Note that we leverage the commonly adopted metric of Intersection over Union (IoU) and the Intersection over Foreground (IoF) ($\text{IoF} = \frac{\text{Area of Overlap}}{\text{Area of Foreground}}$) to measure the relative position. Specifically, we define the detection proposal as {\em sub-text} if its $\text{IoU}_{max} \in (0.1, 0.5)$ (i.e., the {\em sub-text} only covers the fragments of ground truth) and the $\text{IoF}_{max} \in (\beta, 1.0)$ (i.e., most parts of the {\em sub-text} are covered by the ground truth) simultaneously. The detection proposal is defined as {\em full-text} if $\text{IoU}_{max} \in (0.5, 1.0)$. Here we set $\beta$ as 0.7, and will elaborate its impact in Section~\ref{sec.analysis}.

\begin{figure*}[t]
 \centering
 \vspace{-0.1in}
 \includegraphics[width=0.9\linewidth]{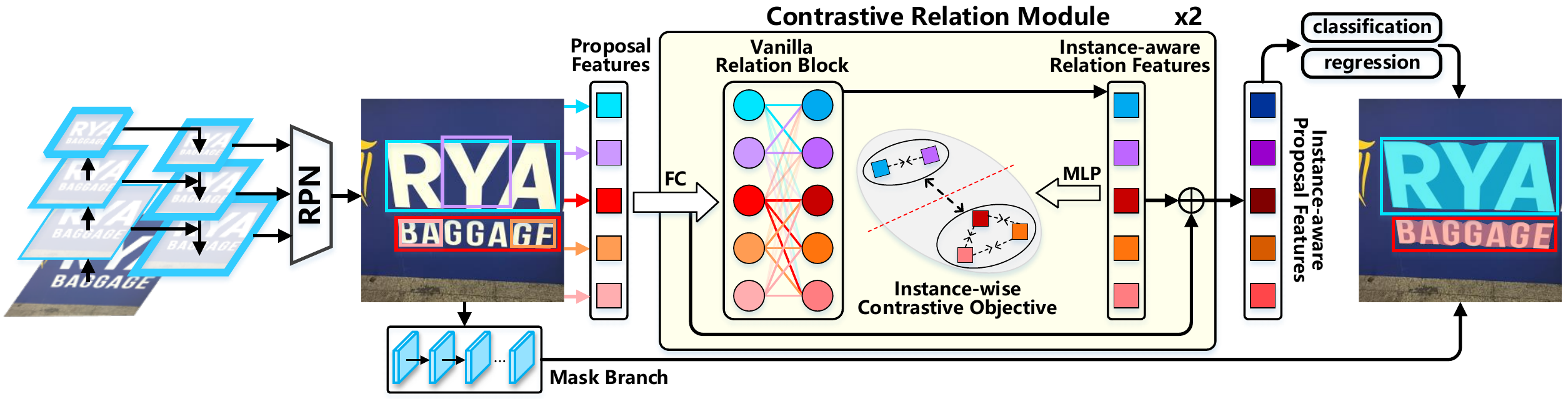}
 \vspace{-0.2in}
 \caption{\small An overview of CORE-Text framework that integrates Mask R-CNN with our CORE module. The input image is first fed into the Feature Pyramid Network (FPN) backbone, and a set of text proposals are produced by Region Proposal Networks (RPN). Then, CORE module performs relational reasoning among all text proposal features. The process of relational reasoning is further guided with instance-wise contrastive objective, that encourages instance-level \emph{sub-text} discrimination. In this way, CORE module learns instance-aware proposal features, which are leveraged to facilitate the classification and regression of text proposals in {\em box} branch of Mask R-CNN. For scene text with arbitrary shape, the {\em mask} branch in Mask R-CNN is additionally utilized to achieve the final segmentation results.}
 \vspace{-0.22in}
 \label{fig:framework}
\end{figure*}

{\bf Frequency.}
To measure the frequency of {\em sub-text} problem, we collect all the bad cases based on ICDAR 2017 MLT {\em val} set. Under a moderate evaluation setting ($\text{IoU} \geq 0.5$), the proportion of {\em sub-text} problem in bad cases is 24.2\%. In addition, we find that more strict evaluation setting will result in more {\em sub-texts}, e.g. the ratio of {\em sub-text} is increased to 49.1\% under $\text{IoU} \geq 0.8$.

{\bf Performance Upper-bound.}
Since the \emph{sub-text} problem accounts for a large proportion of bad cases in existing benchmark, here we investigate the performance upper-bound by fully eliminating {\em sub-texts}. Specifically, for each detected {\em sub-text}, we first measure its IoUs against all ground truths. Next, the detected {\em sub-text} is replaced by the ground truth with the largest IoU for evaluation and the obtained performance is thus treated as the upper-bound. As shown in Figure~\ref{fig:sub-text}(b), after fully eliminating {\em sub-texts}, the Hmean of our base model (Mask R-CNN) is increased by 6\% under moderate evaluation setting ($\text{IoU} \geq 0.5$). Furthermore, under a higher IoU threshold, the larger performance improvement is attained. The results show that there is still much room for improvement in scene text detection, especially for addressing the \emph{sub-text} problem under more strict evaluation.

\section{Approach}
We design a universal module, named COntrastive RElation (CORE), that mitigates the {\em sub-text} problem by jointly performing relational reasoning and instance-level {\em sub-text} discrimination. The CORE module can be further integrated into any two-stage text detector (e.g., Mask R-CNN here) to improve scene text detection. We name the whole text detector as CORE-Text, and Figure~\ref{fig:framework} depicts the detailed architecture.

\subsection{Vanilla Relation Block }
We first provide a brief review of vanilla relation block~\cite{det:rel-net}, which is commonly adopted in generic object detection for relational reasoning among region proposals. Formally, given the input $N$ proposals $\{\text{f}_i^A, \text{f}_i^G\}_{i=1}^N$ ($\text{f}^A$: appearance feature; $\text{f}^G$: geometric feature), vanilla relation block achieves the relation-augmented representation of each proposal by refining appearance feature $\text{f}^A_i$ with $N_r$ learnt relation features as
\begin{eqnarray}
\text{f}_i^A = \text{f}_i^A + \text{Concat}[\{f^{R_m}_i\}_{m=1}^{N_r}].
\label{eq.1}
\end{eqnarray}
Here the $m$-th relation feature $f^{R_m}_i$ of the $i$-th proposal is calculated as the weighted sum of appearance features from other proposals:
\begin{eqnarray}
\text{f}_i^{R_m} = \sum_j w^{m}_{ij} \cdot (W^{V_m} \cdot f^{A}_i),
\end{eqnarray}
where the relation weight $w^{m}_{ij}$ represents the pairwise relation between two proposals based on their appearance and geometric features, and $W^{V_m}$ indicates the transformation matrix. Accordingly, vanilla relation block strengthens proposal representations via relational reasoning that exploits the contextual information among region proposals.

\subsection{Contrastive Relation Module}
The vanilla relation block constructs {\em fully-connected} relations among all region proposals and performs relational reasoning in a self-attention manner. This way apparently leaves the contextual information at instance-level not fully explored, in view that the text proposals (both {\em sub-text} and {\em full-text}) belonging to the same text instance should share the inherently similar semantics. Therefore, we devise the COntrastive RElation (CORE) module by remolding the vanilla relation block with an additional instance-wise contrastive objective. The spirit behind is to guide the relational reasoning with instance-level {\em sub-text} discrimination in a contrastive manner, and thus learn the instance-aware representations of text proposals to mitigate the {\em sub-text} problem. Technically, given the text proposals produced by RPN, CORE module first utilizes vanilla relation block to trigger the relational reasoning that learns relation features of text proposals. The learning of relation features is further guided with instance-wise contrastive objective to enrich relation features with instance-level information. After that, the learnt instance-aware relation features are aggregated with the primary input proposal features via a shortcut connection, leading to the instance-aware features of all text proposals. The instance-aware proposal features are finally fed into the classification and regression modules for text instance localization.

\noindent
{\bf Instance-wise Contrastive Objective.} Traditional contrastive learning~\cite{cl:dr,cai2020joint,yao2020seco} targets for learning feature embedding by attracting positives (semantically similar samples) while repelling negatives (semantically dissimilar samples). The common contrastive objective is InfoNCE~\cite{cl:infonce}, which frames contrastive learning as a classification problem:
\begin{eqnarray}
\L_{\text{CL}}(q,k) = -\text{log}\frac{\text{exp}(q\cdot k^+ / \tau)}{\text{exp}(q\cdot k^+ / \tau) + \sum_{j=1}^K\text{exp}(q\cdot k_j^- / \tau)},
\end{eqnarray}
where $\{q, k^+\}$ is a positive pair, $\{q, k^-_j\}$ is a negative pair, $K$ is the number of negative samples, and $\tau$ is temperature parameter.
Taking the inspiration from contrastive learning, we derive a particular form of loss, i.e., instance-wise contrastive objective, to penalize incompatibility of each text proposal pair. That is to maximize the agreement of different proposals of same text instance, while minimize the agreement of proposals derived from different instances. Formally, conditioned on all the input text proposals from RPN, we first take the $N$ relation features of ground truth proposals as $\{q_i\}_{i=1}^N$. For each query $q_i$, the corresponding positive samples $\{k_{i,m}^+\}_{m=1}^M$ are thus defined as the relation features of both {\em sub-text} and {\em full-text} proposals belonging to the same text instance of $q_i$. Instead, all the relation features of {\em sub-text}, {\em full-text}, and ground truth proposals belonging to different text instances are taken as the negative samples $\{k_{i,j}^-\}_{j=1}^K$. Note that we additionally involve a 2-layer MLP plus ReLU (hidden layer size: 1,024) to transform relation features into a 128-dimensional embedding space in contrastive learning. These output vectors are normalized via a L2-norm layer. Therefore, the instance-wise contrastive loss is calculated as
\begin{eqnarray}
\L_{\text{InsCL}}(q,k) = -\frac{1}{NM}\sum_{i=1}^{N}\sum_{m=1}^{M}\L_{\text{CL}}(q_i,\{k_{i,m}^+,k_i^-\}).
\end{eqnarray}

\subsection{Overall Objective}

Recall that our CORE module is a unified text proposal refiner, it is feasible to plug CORE module into any two-stage text detector for scene text detection. We next present the overall objective of our  CORE-Text by integrating CORE module into Mask R-CNN \cite{ins:mask-rcnn}, which consists of RPN for producing proposal features, {\em box} branch for classification and regression task, and {\em mask} branch for binary segmentation.

Following the multi-task learning paradigm in Mask R-CNN, the overall objective of our CORE-Text is calculated as the integration of RPN loss $\L_{\text{rpn}}$, classification loss $\L_{\text{cls}}$, regression loss $\L_{\text{reg}}$, binary segmentation loss $\L_{\text{mask}}$, and instance-wise contrastive loss $\L_{\text{InsCL}}$:
\begin{eqnarray}
\L = \L_{\text{rpn}}+\L_{\text{cls}} +\L_{\text{reg}} +\L_{\text{mask}} +\lambda\L_{\text{InsCL}},
\end{eqnarray}
where the weight $\lambda$ is set as 0.01 in out experiments.
Note that we adopt two-phase strategy for training CORE-Text. At the first phase, we pretrain CORE-Text with RPN loss and instance-wise contrastive loss. In the second phase, the whole architecture is finetuned with the overall objective $\L$.

\section{Experiments}

We empirically evaluate our CORE-Text on four scene text detection benchmarks: ICDAR 2017 MLT~\cite{dataset:2017}, ICDAR 2015~\cite{dataset:2015}, CTW1500~\cite{dataset:ctw1500}, and Total-Text~\cite{dataset:total-text}.

\subsection{Dataset and Experimental Settings}
\noindent
{\bf Dataset.} {\bf ICDAR 2017 MLT} is a popular benchmark with multi-oriented, multi-scripting and multi-lingual scene texts, containing 7,200 {\em train} images, 1,800 {\em val} images, and 9,000 {\em test} images with word-level annotations. {\bf ICDAR 2015} is another multi-oriented scene text detection benchmark that focuses on English texts, and consists of 1,000 {\em train} images and 500 {\em test} images with annotations labeled as word-level quadrangles. {\bf CTW1500} contains curved texts in natural scenes, and includes 1,000 {\em train} images and 500 {\em test} images with text-line level annotations. {\bf Total-Text} consists of 1,255 {\em train} images and 300 {\em test} images with horizontal, multi-oriented and curved texts. The text instances are annotated at word level.

\begin{table*}[t]
\vspace{-0.12in}
\centering\small
\caption{\small Performance comparisons on four standard benchmark {\em test} sets. H, P, and R are short for Hmean, Precision, and Recall, respectively.}
\label{tab:stoa}
\setlength{\tabcolsep}{3.5mm}{
\begin{tabular}{l||c|c|c||c|c|c||c|c|c||c|c|c}
\bottomrule
\multicolumn{1}{c||}{\bf Method} & \multicolumn{3}{c||}{\bf ICDAR 2017 MLT} & \multicolumn{3}{c||}{\bf ICDAR 2015} & \multicolumn{3}{c||}{\bf CTW1500} & \multicolumn{3}{c}{\bf Total-Text} \\ \cline{2-13}
 {} & {\bf H} & {\bf P} & {\bf R} & {\bf H} & {\bf P} & {\bf R} & {\bf H} & {\bf P} & {\bf R} & {\bf H} & {\bf P} & {\bf R} \\ \hline
CTPN~\cite{text:ctpn} & - & - & - & 61.0 & 74.0 & 52.0 & - & - & - & \multicolumn{1}{c|}{-} & \multicolumn{1}{c|}{-} & - \\ \hline
SegLink~\cite{text:seglink} & - & - & - & 75.0 & 73.1 & 76.8 & - & - & - & \multicolumn{1}{c|}{-} & \multicolumn{1}{c|}{-} & - \\ \hline
Xue {\em et al.}~\cite{text:bootstrap} & 66.6 & 73.9 & 60.6 & - & - & - & - & - & - & \multicolumn{1}{c|}{-} & \multicolumn{1}{c|}{-} & - \\ \hline
DRRG~\cite{text:drrg} & 67.3 & 75.0 & 61.0 & 86.6 & 88.5 & 84.7 & 84.5 & 85.9 & 83.0 & \multicolumn{1}{c|}{85.7} & \multicolumn{1}{c|}{86.5} & 84.9 \\ \hline
PSENet~\cite{text:psenet} & 70.8 & 73.7 & 68.2 & 85.7 & 86.9 & 84.5 & 82.2 & 84.8 & 79.7 & 80.9 & \multicolumn{1}{c|}{84.0} & 78.0 \\ \hline
DB~\cite{text:db} & 74.7 & 83.1 & 67.9 & 87.3 & {\bf 91.8} & 83.2 & 83.4 & 86.9 & 80.2 & \multicolumn{1}{c|}{84.7} & \multicolumn{1}{c|}{87.1} & 82.5 \\ \hline
PMTD~\cite{text:pmtd} & 78.5 & 85.2 & 72.8 & 89.3 & 91.3 & 87.4 & - & - & - & \multicolumn{1}{c|}{-} & \multicolumn{1}{c|}{-} & - \\ \toprule \bottomrule
Base & 77.2 & 82.7 & 72.5 & 88.2 & 90.1 & 86.4 & 84.9 & 86.3 & 83.6 & \multicolumn{1}{c|}{85.3} & \multicolumn{1}{c|}{87.4} & 83.3 \\ \hline
Core-Text & {\bf 78.7}  & {\bf 85.3} & {\bf 73.0} & {\bf 89.3} & 91.1 & {\bf 87.5} & {\bf 85.7} & {\bf 87.8} & {\bf 83.7} & \multicolumn{1}{c|}{\bf 86.3} & \multicolumn{1}{c|}{\bf 87.7} & {\bf 85.0} \\ \toprule
\end{tabular}}
\vspace{-0.26in}
\end{table*}

\noindent
{\bf Network Setups.} We adopt the ImageNet~\cite{cls:imagenet} pretrained ResNet-50~\cite{cls:resnet} with 5-level Feature Pyramid Network~\cite{det:fpn} as the backbone. For {\em prior anchor setting}, we set anchor scale and aspect ratio to 4.82 and \{0.57, 1.10, 1.82, 2.81, 5.54\} by running k-means clustering on {\em train} set bounding boxes. Two stacked CORE modules are utilized to generate the 1,024-d final features for proposal classification and bounding box regression in {\em box} head. Following the default setting of Relation Networks~\cite{det:rel-net}, we set the number of relation features as $N_r=16$ and the dimension of each relation feature is 64. The {\em mask} head contains a four-layer FCN to produce the $28 \times 28 \times 256$ feature map for instance segmentation.

\noindent
{\bf Training Details.} Our model is trained using SGD with 0.9 momentum and 0.0001 weight decay. The batch size is 16. To avoid overfitting, our data augmentation contains: 1) Random horizontal flipping with a probability of 0.5; 2) Random cropping and then resizing to the fixed size $640 \times 640$; 3) Random rotation with an angle range of ($-10^{\circ}$, $10^{\circ}$).

\noindent
{\bf Inference Details.} At inference, we achieve 1,000 proposals by RPN for each testing image. Next, we run the two stacked CORE modules and {\em box} branch on these proposals, followed by Non-Maximum Suppression (NMS) with $\text{IoU} =0.5$. The {\em mask} branch is then applied to the detection boxes, targeting for localizing the scene texts with arbitrary orientations and shapes. Finally, we adopt a mask level NMS with $\text{IoU} =0.2$ to further remove duplicates.

\subsection{Comparisons with State-of-the-Art}
Table~\ref{tab:stoa} summarizes the quantitative results of our CORE-Text on four benchmarks. We compare CORE-Text with several existing state-of-the-art scene text detection techniques.
It is worth noting that we additionally include a degraded version of our CORE-Text (named as \textbf{Base}), which is implemented as the basic Mask R-CNN without CORE module.
Overall, the results across different datasets consistently show that our CORE-Text exhibits better performances than other text detectors over the most metrics. This basically highlights the advantage of performing relational reasoning and instance-level sub-text discrimination for scene text detection.

\begin{table}[t]
\centering\small
\vspace{-0.1in}
\caption{\small Ablation study of CORE module on ICDAR 2017 MLT {\em val} set. VRM: Vanilla Relation Module.}
\label{tab:core}
\setlength{\tabcolsep}{1.9mm}{
\begin{tabular}{l|c|c|c|c}
\bottomrule
Method & Hmean & Precision & Recall & {\em sub-text} number \\ \hline
Base & 80.0 & 82.7 & 77.4 & 1190 \\ \hline
Base + VRM & 81.1 & 85.2 & 77.4 & 923 \\ \hline
Base + CORE & {\bf 82.1} & {\bf 87.1} & {\bf 77.7} & {\bf 754}  \\ \toprule
\end{tabular}}
\vspace{-0.22in}
\end{table}

\noindent{\bf ICDAR 2017 MLT}. At the first training phase, we pretrain our CORE modules with instance-wise contrastive objective for 40 epoch, with the initial learning rate 0.02 annealed by the cosine decay rule. Following the commonly adopted setting in~\cite{text:pmtd}, we train CORE-Text on both ICDAR 2017 MLT {\em train} and {\em val} set for 160 epochs at the second training phase. The learning rate is set to 0.04, which is decreased by one-tenth at 80-th and 128-th epoch respectively. During inference, we adopt the single scale testing strategy and resize the long side of each image to 1,920. Our CORE-Text achieves 78.7\% Hmean, which makes the absolute improvement over the Base model by 1.5\%.

\noindent{\bf ICDAR 2015}. As in \cite{text:pmtd}, we initialize CORE-Text with the ICDAR 2017 MLT pretrained model and further finetune the model with 40 epochs over ICDAR 2015 {\em train} set. The learning rate is set to 0.002 and decays one-tenth at 20-th epoch. At inference, the long side of images is resized to 1,920. Finally, CORE-Text achieves 1.1\% higher Hmean than Base model.

\noindent{\bf CTW1500 \& Total-Text}. To fully verify the generalizability of our CORE-Text, we further evaluate CORE-Text on two challenging benchmarks with curved and multi-oriented texts (i.e., CTW1500 and Total-Text). As in the training settings on ICDAR 2015, we finetune CORE-Text with 40 epochs on each benchmark. The long side of images is resized to 640 and 1,280 on CTW1500 and Total-Text, respectively.
Similar to the observations on ICDAR 2015, our CORE-Text consistently outperforms the Base model by 0.8\% and 1.0\% in Hmean on CTW1500 and Total-Text.

\subsection{Experimental Analysis}
\label{sec.analysis}
Then, we conduct ablation study to verify the effectiveness of our CORE module, and further analyze the impact of several hyper-parameters in CORE-Text. Note that all discussions here are based on ICDAR 2017 MLT {\em train} and {\em val} set. Specifically, we train CORE-Text on ICDAR 2017 MLT {\em train} set with 40 epochs, and the initial learning rate is set as 0.04 (decreased by 10 at 20-th and 32-th epoch respectively). The final results are reported over the 1,800 {\em val} images.

\noindent{\bf Ablation study.} To examine the impact of each design in CORE module, we conduct ablation study by comparing different variants of CORE-Text in Table~\ref{tab:core}. We start from the Base model which is a degraded version of CORE-Text without CORE module. Next, we extend Base model by involving the Vanilla Relation Module (VRM) to trigger relational reasoning among proposals in a self-supervised manner, which achieves better performances and meanwhile reduces the number of \emph{sub-text} bad cases. The results basically demonstrate the effectiveness of relational reasoning in VRM. In addition, the integration of both relational reasoning and instance-level \emph{sub-text} discrimination, i.e., our CORE module, obtains the highest performances in terms of all the three metrics and further reduces the \emph{sub-text} number. The performance gains validate the merit of guiding relational with instance-wise contrastive objective for scene text detection.

\noindent{\bf Impact of $\beta$ in \emph{sub-text} definition.} Table~\ref{tab:beta} shows the results by varying $\beta$ in the range of [0.5, 0.9]. The Hmean scores only change between 81.7\% and 82.1\%, which practically eases the selection for the optimal $\beta$ in \emph{sub-text} definition.

\noindent{\bf Impact of temperature $\tau$.}  The temperature $\tau$ controls the flatness of softmax function in contrastive loss. As shown in Table~\ref{tab:tau}, the best performance is attained when $\tau$ is set as 0.2. Note that the training loss fails to converge when $\tau=0.01$.

\begin{figure}[t]
\centering
\vspace{-0.02in}
	\includegraphics[width=0.95\linewidth]{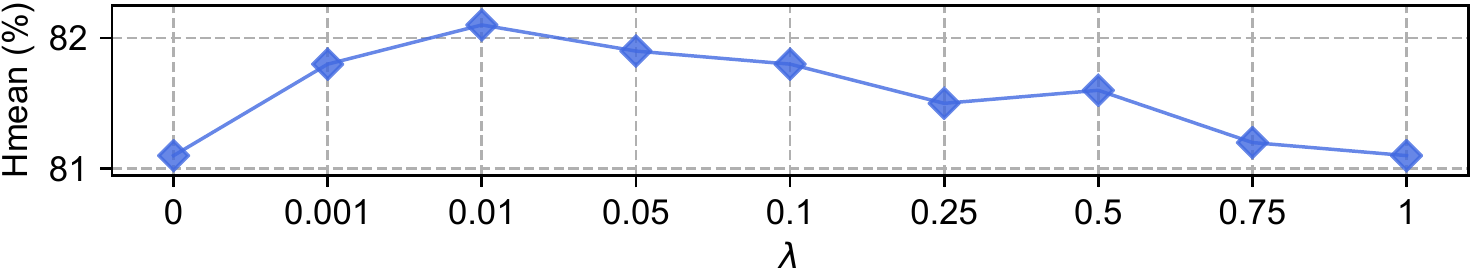}
\vspace{-0.25in}
	\caption{\small Impact of instance-wise contrastive loss weight $\lambda$.}
\vspace{-0.25in}
\label{fig.lambda}
\end{figure}

\noindent{\bf Impact of instance-wise contrastive loss weight $\lambda$.} Figure \ref{fig.lambda} depicts the performance curve of Hmean when $\lambda$ varies within [0, 1]. In the extreme case of $\lambda=0$, no instance-level \emph{sub-text} discrimination is performed and the CORE module degenerates to vanilla relation module. The best Hmean score is achieved when $\lambda=0.01$.  This again demonstrates that it is reasonable to exploit both relational reasoning and instance-level \emph{sub-text} discrimination for boosting scene text detection.

\begin{table}[t]
\vspace{-0.15in}\small
\caption{\small Impact of hyper-parameter $\beta$ in {\em sub-text} definition on ICDAR 2017 MLT {\em val} set.}
\label{tab:beta}
\centering
\setlength{\tabcolsep}{3.6mm}{
\begin{tabular}{c|c|c|c|c|c}
\bottomrule
$\beta$ & 0.5 & 0.6 & 0.7 & 0.8 & 0.9 \\ \hline
Hmean (\%) & 82.0 & 81.7 & {\bf 82.1} & 81.8 & 82.0 \\ \toprule
\end{tabular}}
\vspace{-0.2in}
\end{table}

\begin{table}[t]
\vspace{-0.1in}\small
\caption{\small Impact of temperature $\tau$ in contrastive loss on ICDAR 2017 MLT {\em val} set.}
\label{tab:tau}
\centering
\setlength{\tabcolsep}{2mm}{
\begin{tabular}{c|c|c|c|c|c|c|c}
\bottomrule
$\tau$ & 0.01 & 0.05 & 0.1 & 0.2 & 0.3 & 0.4 & 0.5 \\ \hline
Hmean (\%) & - & 80.7 & 82.0 & {\bf 82.1} & 81.7 & 81.4 & 81.6 \\ \toprule
\end{tabular}}
\vspace{-0.26in}
\end{table}

\section{Conclusions}

In this paper, we investigate the {\em sub-text} problem in scene text detection task and present a novel COntrastive RElation (CORE) module to alleviate this issue. Particularly, unlike existing methods that tackle {\em sub-text} problem in a two-phase fashion, our CORE module jointly localizes and associate text proposals (\emph{sub-texts} and \emph{full-texts} from multiple instances) to boost scene text detection. To materialize our idea, we remold the vanilla relation block by additionally involving an instance-wise contrastive objective to guide the process of relational reasoning among text proposals. Such design naturally enables a joint learning of relational reasoning and instance-level {\em sub-text} discrimination in a contrastive manner, leading to instance-aware representations of text proposals. Furthermore, we devise a novel text detector (i.e., CORE-Text) that integrates CORE module into the generic object detector (Mask R-CNN). Extensive experiments conducted on four benchmarks demonstrate the efficacy of CORE-Text.

{\small
\bibliographystyle{IEEEbib}
\bibliography{icme2021template}
}

\end{document}